\documentclass[10pt,twocolumn,letterpaper]{article}

\usepackage{cvpr}              % To produce the CAMERA-READY version
\usepackage{graphicx}
\usepackage{amsmath}
\usepackage{amssymb}
\usepackage{booktabs}

\usepackage{multirow}
\usepackage{hhline}
\usepackage[accsupp]{axessibility}  % Improves PDF readability for those with disabilities.

\usepackage[pagebackref,breaklinks,colorlinks]{hyperref}

\usepackage[capitalize]{cleveref}
\crefname{section}{Sec.}{Secs.}
\Crefname{section}{Section}{Sections}
\Crefname{table}{Table}{Tables}
\crefname{table}{Tab.}{Tabs.}

\def\cvprPaperID{2} % *** Enter the CVPR Paper ID here
\def\confName{CVSports}
\def\confYear{2022}

\begin{document}

%%%%%%%%% TITLE
\title{Efficient tracking of team sport players with few game-specific annotations}

\author{Adrien Maglo \hspace{2cm} Astrid Orcesi \hspace{2cm} Quoc-Cuong Pham\\
Université Paris-Saclay, CEA, List, F-91120, Palaiseau, France\\
{\tt\small \{adrien.maglo, astrid.orcesi, quoc-cuong.pham\}@cea.fr}
}

\maketitle

\newcommand{\am}[1]{\textcolor[rgb]{1,0,0}{#1}}
\newcommand{\ao}[1]{\textcolor[rgb]{0,0,1}{#1}}

%%%%%%%%% ABSTRACT

\begin{abstract}

One of the requirements for team sports analysis is to track and recognize players. Many tracking and re-identification methods have been proposed in the context of video surveillance. They show very convincing results when tested on public datasets such as the MOT challenge. However, the performance of these methods are not as satisfactory when applied to player tracking. Indeed, in addition to moving very quickly and often being occluded, the players wear the same jersey, which makes the task of re-identification very complex.
Some recent tracking methods have been developed more specifically for the team sport context. Due to the lack of public data, these methods use private datasets that make impossible a comparison with them.
In this paper, we propose a new generic method to track team sport players during a full game thanks to few human annotations collected via a semi-interactive system.
Non-ambiguous tracklets and their appearance features are automatically generated with a detection and a re-identification network both pre-trained on public datasets. Then an incremental learning mechanism trains a Transformer to classify identities using few game-specific human annotations. Finally, tracklets are linked by an association algorithm.
We demonstrate the efficiency of our approach on a challenging rugby sevens dataset. To overcome the lack of public sports tracking dataset, we publicly release this dataset at \url{https://kalisteo.cea.fr/index.php/free-resources/}.
We also show that our method is able to track rugby sevens players during a full match, if they are observable at a minimal resolution, with the annotation of only 6 few seconds length tracklets per player.

\end{abstract}

%%%%%%%%% BODY TEXT
\section{Introduction}

Player tracking in team sports consists in detecting and identifying the players in video sequences. It is a necessary task to automate the generation of individual statistics such as ball possession, field position or involvement in play sequences.
Player tracking in team sports such as rugby is however a challenging task. Rugby is a sport of physical contact where player occlusions are very frequent on camera during rucks, tackles and scrums. The players can also adopt a wide range of body postures from sprinting to laying on the ground in a foetal position. Players from the same team share a very similar appearance since they wear the same jerseys. Moreover, the number of pixels in which the players are visible is often limited in the case of a TV stream (sometimes with a height below 150 pixels). This prevents the access to fine identification details.

\begin{figure}
\begin{center}
\includegraphics{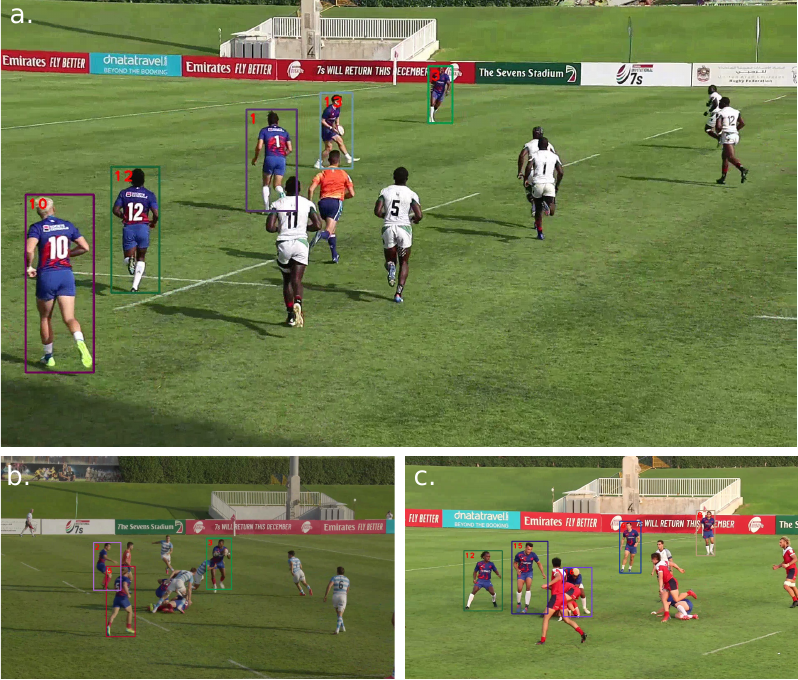}
\end{center}
   \caption{Tracking French players (blue jerseys) in our rugby sevens dataset: a. France / Kenya extract, b. Argentina / France extract, c. France / Chile extract.}
\label{fig:captures}
\end{figure}

Player tracking is a specific Multi-Object Tracking (MOT) problem. MOT has been widely studied in the literature. Security applications have lead to the development of many people tracking approaches. Offline methods use all the frames of the input video sequence to optimize the generation of tracks while online methods target real-time applications by relying only on the current and previous frames to generate the tracks.
The most recent frameworks achieve the best performance using deep neural network architectures.
The availability of large public datasets and challenges such as the MOT challenge \cite{milan2016mot} allows to fairly train and compare the various approaches.

Some recent people tracking methods have also been proposed for the specific context of team sports: soccer \cite{zhang2020multi, hurault2020self}, basketball \cite{lu2013learning} and hockey \cite{vats2021player}.
These methods often use private datasets specific to their studied sport to get competitive results evaluated on short video clips extracted from a match.
Game-specific annotations are required to train a player tracking and identification system to adapt to the player identities and the context of a game. The number of such annotations is an important factor that will determine the success of using such a system in real world scenario.
Little focus was made in previous work on the practicability of this annotation process. Consequently, we propose an incremental learning approach to identify players with very few game-specific annotations.
Our method is offline: it tracks and identifies players once the game has been completed. It benefits from the closed gallery re-identification (re-ID) hypothesis as, contrary to video surveillance, the number of players is known and limited. Since our method does not use any sport-specific knowledge, it can be applied to any team sport.

Our annotation process consists in several steps. Bounding boxes around all the persons in the frames are first extracted from the input video to generate non-ambiguous tracklets. A tracklet is the uninterrupted sequence of the bounding box images of a single player. Tracklets can have a variable length since a player can enter or leave the camera field of view or be occluded by an other player. At this stage, the user provides few annotations per player to train the tracklet re-ID network. Finally, the obtained tracklet classification scores or appearance features feed an algorithm that look for an optimal association between tracklets and identities.

The contributions of this paper are the following:

We tackle the sparsity of training data in team sport contexts by leveraging generic detection and re-ID datasets. The detection network is only learned on a public dataset. The re-ID network is pretrained on a video surveillance public dataset.

We propose a new architecture based on a Transformer network \cite{vaswani2017attention} to classify and generate tracklet appearance features. An incremental learning mechanism using few user interactions trains this model and strengthen the re-ID performances throughout the whole annotation process.

Some datasets have been proposed for basketball \cite{delannay2009detection} and soccer \cite{dorazio2009semi} player tracking with multiple static cameras. However, although Deliege et al. \cite{deliege2021soccernet} are extending their SoccerNet dataset to tracking and re-ID, no dataset with a moving point of view has been made available.
We publicly release our rugby sevens tracking dataset composed of single-view videos that can pan, tilt or zoom to follow the action. It is one of the most challenging team sport for tracking on which no approach was tested.

We demonstrate the efficiency of our approach on our dataset. On a full game, it can achieve up to 67.9\% detection and identity classification recall when the players are sufficiently visible with only 6 annotations per player.

The paper is organized as follows:
Section~\ref{sec:sota} introduces Related Work. Our method is described in Section~\ref{sec:method}. Finally, Section~\ref{sec:results} provides our results on our challenging rugby sevens dataset, compares them to state-of-the-art methods and analyzes them in an ablative study.

\section{Related Work}
\label{sec:sota}

\subsection{Multiple people tracking}

Two categories of MOT algorithms can be distinguished.

\textbf{Offline methods} leverage the full sequence of images to globally optimize the generated tracks with a graph paradigm. The vertices are the detections on each frame and the edges are the connections between detections that form tracks.
Thus, Zhang et al. \cite{zhang2008} uses a minimum cost flow iterative algorithm that models long term occlusions.
The approach described by Berclaz et al. \cite{berclaz2011} takes only an occupancy map of detection as input and applies a k-shortest path algorithm on the flows.
More recently, Brasó and Leal-Taix \cite{braso2020learning} proposed a fully differentiable network that learns both the appearance and geometrical feature extraction as well as the detection association to generate tracks. Hornakova et al. \cite{hornakova2020lifted} use lifted edges to model long term interactions and generate the optimized solution with a linear programming relaxation.

\textbf{Online methods} only use the current and past frames to associate new detections with tracks. They have raised more interest in the literature as they fit real time scenario.
Thus, the SORT algorithm \cite{bewley2016} uses a Faster R-CNN \cite{ren2015} person detector. Then, a Kalman filter \cite{kalman1960new} predicts the future positions of each track. The Intersection-Over-Union (IOU) between these predictions and the detected bounding boxes are used as inputs of an Hungarian algorithm that matches the detection with the tracks.
ByteTrack \cite{zhang2021bytetrack} achieves state of the art tracking performance with a two steps association algorithm: the first step focuses on high confidence detections while the second step deals with the low confidence ones.
The Deep SORT algorithm \cite{wojke2017simple} adds a re-ID network to extract the visual appearance of each person.
The input data of the Hungarian algorithm becomes a combination of a Manaholis distance as the spatial term and a cosine distance between the re-ID vectors as the appearance term.

Using distinct networks for detection and re-ID has the advantage of separating the two tasks that may have opposite objectives.
The detection task aims at learning common features to recognize humans while the re-ID task aims at learning distinctive features of each individual.
However this may cause scalability issues as each detected bounding box must be independently processed by the re-ID network.
Single-shot methods were therefore proposed to generate the bounding boxes coordinates and re-ID vectors with a single network.
Thus, Track-RCNN \cite{voigtlaender2019} uses a common backbone with specific heads for each task. FairMOT \cite{zhang2020fairmot} achieves better tracking performance by focusing only on the detection and re-ID tasks. Meinhard et al. \cite{meinhardt2021trackformer} use a Transformer architecture.

Applying traditional MOT to team sport players usually leads to many ID switches.
Each time a player leaves the vision field or is occluded too much time, a new identity is generated at reappearance. This prevents the reliable generation of individual statistics (see section \ref{comparison_generic_tracking}).

\subsection{Multiple team sport player tracking and re-identification}

\subsubsection{Tracking}

Some tracking methods have been proposed for the context of team sports. For soccer, many approaches performed tracking by first extracting the field regions \cite{manafifard2017survey, khatoonabadi2009automatic, baysal2015sentioscope, liu2009automatic, d2009investigation, xing2010multiple}.
In the method of Liu et al. \cite{liu2009automatic}, an unsupervised clustering algorithm classifies the players among four classes (two teams, referee or outlier). The tracking is formulated as a Markov chain Monte Carlo data association.
D'Orazio et al. \cite{d2009investigation} classify each player with an unsupervised clustering algorithm.
The tracking takes as input geometrical and motion information. It is based on a set of logical rules with a merge split strategy.
In Xing et al. \cite{xing2010multiple}, the observation model of each player is composed of the color histogram in the cloth regions, the size and the motion. The tracking is formulated as particle filtering.
Theagarajan and Bhanu \cite{theagarajan2020automated} used a YOLOv2 \cite{redmon2016you} network detector and a DeepSORT tracker \cite{wojke2017simple} to identify the player controlling the ball.

All the previous approaches do not build individual appearance signatures per player identities.
If a player leaves the camera field of view and re-enter later, he/she will be considered as a new person. This prevents the generation of individual statistics.

\subsubsection{Re-identification}

Jersey number recognition has been studied in the literature to identify team sport players. Ye et al. \cite{ye2005jersey} developed a method based on Zernike moments features \cite{khotanzad1990invariant}. Gerke et al. \cite{gerke2015soccer} were the first to use a convolutional neural network to classify jersey numbers from bounding box images of players. It was later combined to spatial constellation features to identify soccer players \cite{gerke2017soccer}. To ease the recognition of distorted jersey numbers, Li et al. \cite{li2018jersey} trained a branch of their network to correct the jersey number deformation before the classification. Liu and Bhanu \cite{liu2019pose} enabled jersey number recognition only in the relevant zones by detecting body keypoints. For hockey player identification, Chan et al. \cite{chan2021player} used a ResNet + LSTM network \cite{he2016deep, hochreiter1997long} on tracklet images to extract jersey numbers. 

When a single view is available, as in our rugby sevens dataset, jersey numbers are often not visible, partially visible or distorted. Besides, to our knowledge, there is no publicly available training dataset for team sport jersey number recognition.
A solution can therefore be to use appearances to re-identify players.
Teket and Yetik \cite{teket2020fast} proposed a framework to identify the player responsible for a basketball shot. Their re-ID network, based on MobileNetV2 \cite{sandler2018mobilenetv2}, is trained with a triplet loss formulation.
The framework described by Senocak et al. \cite{senocak2018part} combines part-based features and multiscale global features to generate basketball player signatures. Both approaches are based, as ours, on the hypothesis of a closed gallery however they use a private dataset to train their model which makes comparisons impossible.

\subsubsection{Tracking with re-identification}

Several methods tracks players by using re-ID features \cite{lu2013learning, zhang2020multi, yang2021multi, hurault2020self, vats2021player}.
Lu et al. \cite{lu2013learning} use DPM \cite{felzenszwalb2008discriminatively} to detect basketball players. Local features and RGB color histograms are extracted on players for the re-ID.
Zhang et al. \cite{zhang2020multi} proposed a multi-camera tracker that locates basketball players on a grid based on a K-shortest paths algorithm \cite{berclaz2011}. Players are detected and segmented with a network based on Mask R-CNN \cite{he2017mask}. Re-ID features are computed thanks to the team classification, jersey number recognition and a pose-guided feature embedding.
To track soccer players, Yang et al. \cite{yang2021multi} iteratively reduced the location and identification errors generated by the previous approach by creating a bayesian model that is optimized to best fit input pixel level segmentation and identification.
Hurault et al. \cite{hurault2020self} use a single network with a Faster R-CNN backbone \cite{ren2015} to detect small soccer players and extract re-ID features. Kong et al. \cite{kong2021online} mix player appearance, posture and motion criteria to match new detections with existing tracks.
Vats et al. \cite{vats2021player} use a Faster R-CNN network \cite{ren2015} to detect hockey players and a batch method for tracking \cite{braso2020learning}. Specific ResNet-18 networks \cite{he2016deep} are used to identify the player teams and jersey numbers.

Most of the approaches presented here \cite{lu2013learning, zhang2020multi, yang2021multi, vats2021player} train their re-ID or jersey number recognition model with a private dataset.

\subsubsection{Minimizing the number of annotations}

To our knowledge, few previous work focus on the minimization of the game-specific training annotations for re-ID.
For example, Lu et al. \cite{lu2013learning} used a mere 200 labels for every players in a team with their semi-supervised approach.
Senocak et al. \cite{senocak2018part} use 2500 cropped images for each player to train their re-ID network.
Teket and Yetik \cite{teket2020fast} use a training dataset that contains 30 to 1000 images per player.
In this paper, by asking the user to annotate tracklets, we aim to demonstrate that it is possible to produce meaningful player re-ID results for a rugby sevens full game with only 6 annotations per player.

\section{Proposed method}
\label{sec:method}

\subsection{Overview}

\begin{figure}
\begin{center}
\includegraphics{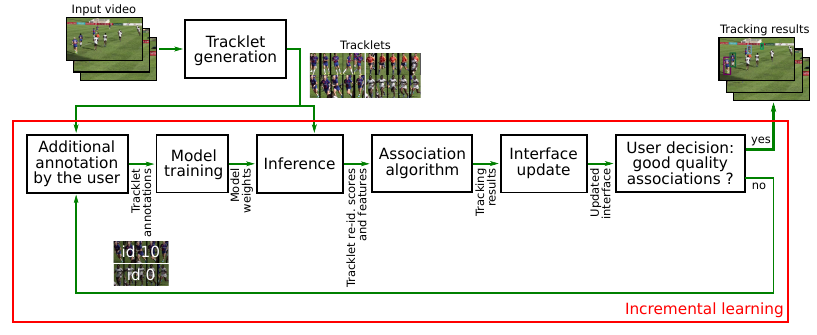}
\end{center}
   \caption{Incremental learning of tracklet classification. The user provides annotations to train the model to correctly classify the tracklets to a player identity.}
\label{fig:process}
\end{figure}

We propose a new method to track the \(N_p\) players of a team in a video with a single moving view of a game. The first step of our method generates \( N_{t} \) tracklets we qualify as non-ambiguous because they contain a single identity. For this purpose, bounding boxes around persons are detected and associated across frames automatically.
The user can then provide few identity annotations to some of the generated tracklets thanks to a dedicated interface show on Figure \ref{fig:interface_capture}.
The tracklet re-ID network can then be trained with these annotations.
Once the model is trained, classification scores and re-ID features are generated for all the tracklets.
This data feeds an algorithm that matches every tracklet to an identity.
Once the annotation interface has been updated, the user can then decide to add more annotations to correct the wrong classifications or to stop this incremental learning mechanism if she/he is satisfied by the results.
The whole process is depicted on Figure \ref{fig:process}.

\subsection{Tracklet generation}
\label{sec:tracklet_generation}

Non-ambiguous tracklets are generated with a tracking by detection paradigm. A Faster R-CNN network \cite{ren2015} with a ResNet-50 backbone \cite{he2016deep} trained on the COCO dataset \cite{lin2014microsoft} detects all the persons in the video frames. This detector is a well-known model used in several recent work \cite{hurault2020self, vats2021player}. To generate the tracklets, we use the simple and classic approach described in \cite{bewley2016}. Bounding boxes between the previous and the current frames are associated by bipartite matching with an Hungarian algorithm \cite{kuhn1955hungarian}. This matching is performed with a single IoU criteria since the player appearances are later taken into account by our tracklet re-ID model. We also use a Kalman filter \cite{kalman1960new} to predict the position of an existing track in the current frame.

Each generated tracklet will be later matched to a single identity. We therefore want to avoid as much as possible identity switches inside tracklets.
When a tracklet partially occludes an other one, bipartite matching may generate a wrong association. Our algorithm therefore splits the tracklets that intersect since they are considered as ambiguous. If at the current frame, two tracklet bounding boxes have an IoU above a threshold \(\mu = 0.5\) these tracklets are terminated and new ones are created. We also filter out tracklets that have a length inferior to \( l_{min} \). We indeed consider that they may also be ambiguous by containing several identities in their images. Besides, they do not provide enough diverse data to the tracklet re-ID model.

\subsection{Incremental learning tracklet classification}

\begin{figure}
\begin{center}
\includegraphics[scale=0.53]{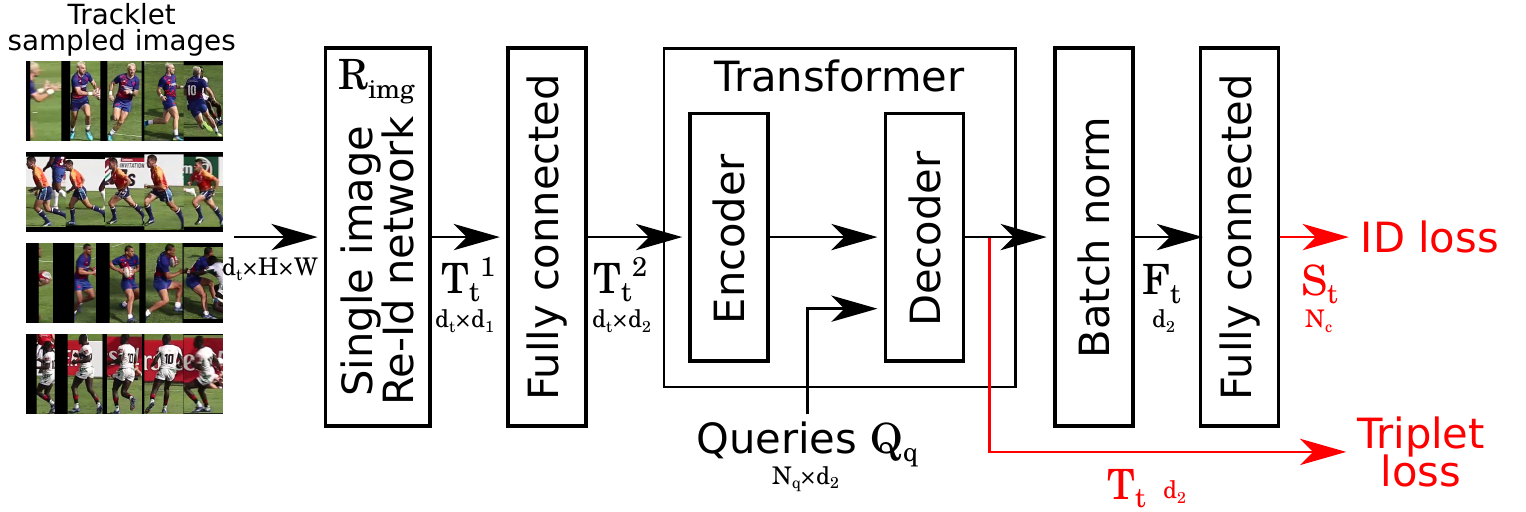}
\end{center}
   \caption{Architecture of the tracklet classification network. \( R_{img} \) extracts re-ID features \( T^1_t \) from the tracklet images. They are combined by the transformer to generate a single tracklet re-ID vector \( F_t \). The model is trained by ID loss and triplet loss.}
\label{fig:archi}
\end{figure}

The aim of our system is to match tracklets to identities with the fewest possible annotations. This process is done through incremental learning since the user can choose to add more training annotations while the quality of the generated tracklet association is not satisfying.
We set the target number of classes \(N_c = 1 + N_p\). The class zero corresponds to all persons we do not want to track (players from the opponent team, referees, public). Our tracklet re-ID model is mainly composed of a single image re-ID network \( R_{img} \) followed by a Transformer \cite{vaswani2017attention} as illustrated on Figure \ref{fig:archi}.

For \( R_{img} \), we chose the model described by Luo et al. \cite{luo2019bag} for its simplicity. It uses a ResNet-50 backbone \cite{he2016deep} and has been trained on the generic Market1501 dataset \cite{zheng2015scalable}. It takes as input single images at resolution \(H \times W\) and outputs player appearance features at dimension \(d_{1}\).
We regularly sample \(d_{t}\) images from each tracklet and combine their appearance features to obtain the tracklet features tensor \( T^1_t \in \mathbb{R}^{d_{t} \times d_{1}} \).
The feature dimension of \(T^1_t\) is then reduced to \(d_{2}\) by a fully connected layer to obtain \(T^2_t\). This limits the dimension of the features inside the next nodes of our model in order to train it quickly.

The Transformer in our model then combines the re-ID features of the sampled tracklet images \(T^2_t\) to generate a single tracklet re-ID vector \( F_t \). Its cross-attention nodes can learn to focus on the most distinctive features across the tracklet sampled frames.
It takes as input of the encoder \(T^2_t\) and as input of the decoder the \(N_{q}\) queries \(Q_q\). 
Similarly to DETR \cite{carion2020end}, the queries \(Q_q \in \mathbb{R}^{d_{2}}\) are learned embeddings. Each query learns to specialize on some features of the player identities. However, we do not use any input positional encoding because, since our initial variable length tracklets are resampled to fixed length \(d_{t}\), there are no common temporal link between the features.
We found that using 16 encoder layers, one decoder layer and 16 heads in the multi-head attention models was the best set of parameters.
At the output of the decoder, a batch norm layer generates the tracklet features \( F_t \). For the classification, a fully connected layer computes the classification scores \(S_t \in \mathbb{R}^{N_{c}} \). 

Given a tracklet \( t \), the \(N_{qc}\) queries among \(N_{q}\) that gives the highest classification scores are selected for the back-propagation. The optimized loss is defined by
\[
L = L_{ID}(S_t, \hat{S_t}) + \alpha L_{Triplet}(D_{t,p}, D_{t,n}),
\]
where \( L_{ID} \) is the standard cross entropy loss, \( \hat{S_t} \) are the target classification logits,  \( L_{Triplet} \) is the soft-margin triplet loss \cite{hermans2017defense}, \( D_{t,p} \) and \( D_{t,n} \) are feature distances of positive pairs and negative pairs and \( \alpha \) is a binary constant. As described by Luo et al. \cite{luo2019bag}, the idea of combining a classification loss and a triplet loss is to let the model learn more discriminative features \( F_t \in \mathbb{R}^{d_2} \). For the triplet loss, we use a batch hard strategy that finds the hardest positive and negative samples.

Once the model has been trained, all tracklets are processed by the model at inference stage to compute the tracklet classification scores \( S_t \) and features \( F_t \).

\subsection{Association algorithms}

With generated scores \(S_t\) and the features \(F_t\), we have the needed data to match tracklets to player identities by using an association algorithm. Two alternative methods are investigated.

\subsubsection{Iterative association}

An iteration of the association algorithm consists in selecting the highest score in the matrix of all tracklet scores \(S_t\). The highest score represents a matching between the tracklet \(t\) and the identity \(i\). The algorithm then checks that \(t\) can be associated to \(i\) by verifying that the tracklets already associated to \(i\) do not already appear in the frames where \(t\) appears. If the association is possible, \(t\) is added to the list of tracklets associated to \(i\) and a new iteration of the algorithm is run.
When the iterative association is used, we set \( \alpha = 0 \) during the incremental learning to only optimize the classification scores \( S_t \).

\subsubsection{Matrix factorization association}
\label{sec:rnmf}

The second algorithm is inspired from \cite{he2020multi}. The authors describe a multi-camera batch people tracking system that assigns tracklets extracted from different views to identities. The input of the algorithm is a tracklet similarity matrix \(S\) generated with appearance, motion and localization criteria. A Restricted Non-negative Matrix Factorization (RNMF) algorithm optimizes the identity assignment. The association matrix \(A \in \mathbb{R}^{N_{t} \times N_{p}} \) is computed thanks to the iterative updating rule given in \cite{ding2008convex}.
We applied the RNMF algorithm to our single view case with \(S\) as the sum of an appearance term \(\Psi_{app}\) and a localization term \( \Psi_{loc} \). The similarity between two tracklets \(u\) and \(v\) is computed with:
\[S(u,v) = clip(\Psi_{app}(F_u, F_v)) + clip(\Psi_{loc}(B_{ul}, B_{vf})) \]
where \( clip(x) = max(min(x; 1); 0) \).

\(\Psi_{app} \) is defined by equation \ref{eq:psi_app}.
\begin{equation}
\label{eq:psi_app}
\Psi_{app(F_u, F_v)} = 1 - \frac{1}{\eta_{app}} \cdot d(F_u, F_v)
\end{equation}
where \(d(F_u, F_v)\) is the cosine distance between the feature vectors of the two tracklets and \(\eta_{app}\) is the cosine distance threshold above which we consider that \(u\) and \(v\) belongs to two distinct identities.

\(\Psi_{loc} \) is defined by the equation \ref{eq:psi_loc}. \( t_{ul} \) is the end time of the of the first tracklet and \( t_{vf} \) is the start time of the second tracklet. \(B_{ul}\) and \(B_{vf}\) are the corresponding bounding boxes.
\begin{multline}
\label{eq:psi_loc}
    \Psi_{loc}(B_{ul}, B_{vf}) = \\
    \begin{cases}
        (1 + \eta_{loc}) \cdot IoU(B_{ul}, B_{vf}) - \eta_{loc} & \text{if } t_{vf} - t_{ul} \leq  \tau \\
        0,              & \text{otherwise}
    \end{cases}
\end{multline}
where \(\eta_{loc}\) and \( \tau \) are constant numbers.
\(\Psi_{loc} \) aims at giving a high similarity scores to two successive tracklets if \(B_{ul}\) and \(B_{vf}\) have a high IoU.
When the RNMF association is used, we set \( \alpha = 1 \) during the incremental learning.

\section{Experimental Results}
\label{sec:results}

\subsection{Implementation details}

\begin{figure}
\begin{center}
\includegraphics[scale=0.225]{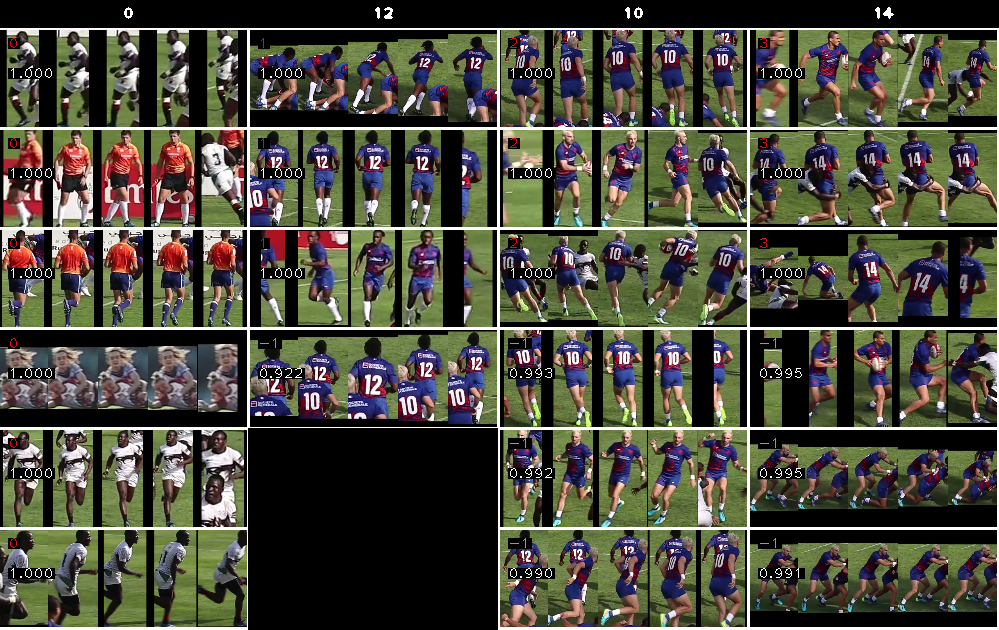}
\end{center}
   \caption{Partial screen capture of our semi-interactive annotation interface. Each cell corresponds to one tracklet. Each column corresponds to one player identity, except the zero column that contains all the persons we do not want to track.}
\label{fig:interface_capture}
\end{figure}

Our system is implemented using the Pytorch framework.
The minimum number of frames of a tracklet \( l_{min} \) is set to 10. All the tracklets are resampled to \(d_t = 10\).
Our re-ID network \cite{luo2019bag} takes as input images of resolution \(H=256\) and \(W=128\). It outputs features at dimension \(d_{1} = 2048\).
Our Transformer network takes input features at \(d_{2} = 128\). The number of input queries \(N_{q}\) is set to 32. They are randomly initialized. The number of queries selected for backpropagation \(N_{qc}\) is set to 4.
It is trained during 120 epochs with an AdamW optimizer, a learning rate of \(9 \times 10^{-5}\), a weight decay of \(10^{-4}\) and a batch size of 4.
The transformer parameters are initialized with Xavier initialization \cite{glorot2010understanding}.
For the linear layer, He initialization \cite{he2015delving} is used.
\(\eta_{app}\) and \(\eta_{loc}\) are experimentally set to 0.35 and 0.43.
The time threshold \(\tau\) for the localization similarity is set to 0.5 seconds.

Our semi-interactive annotation interface, illustrated on Figure \ref{fig:interface_capture}, can run on a laptop GPU (Quadro M2000M). It allows the annotator to generate training data for our model by indicating to which player belongs a tracklet.
The training time represents about 0.8 second per annotation when \(R_{img}\) is frozen and the iterative association is used.

\subsection{Player tracking on rugby sevens samples}
\label{tracking}

\begin{figure*}
\newcommand{\graphScale}{0.344}
\begin{center}
\setlength\tabcolsep{0pt}
\begin{tabular}{cc}
\includegraphics[scale=\graphScale]{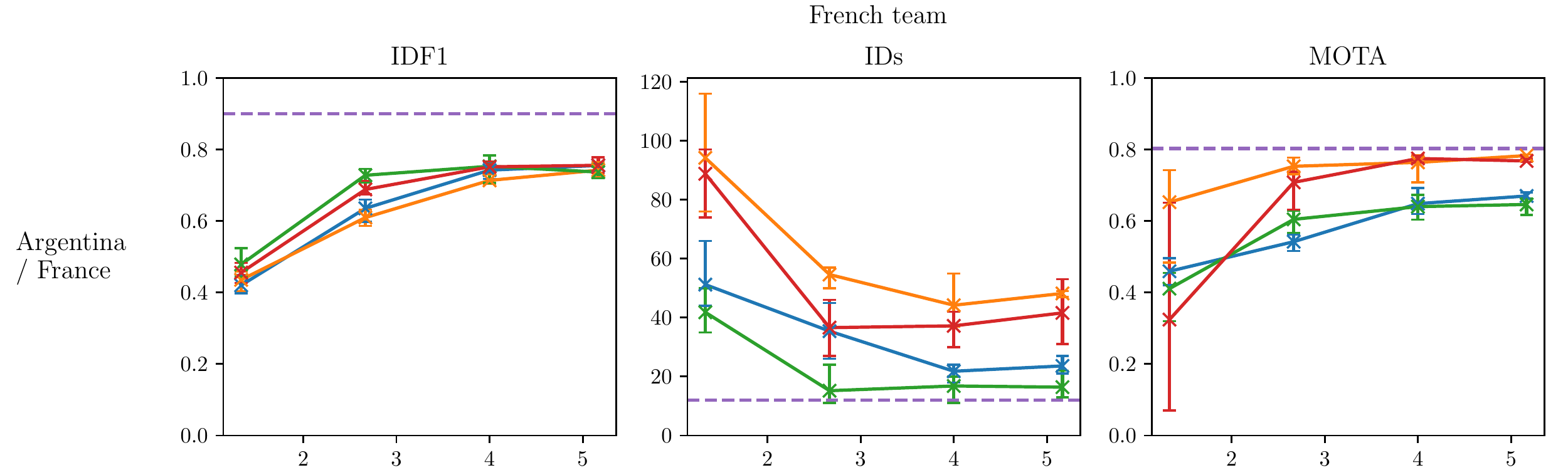} &
\includegraphics[scale=\graphScale]{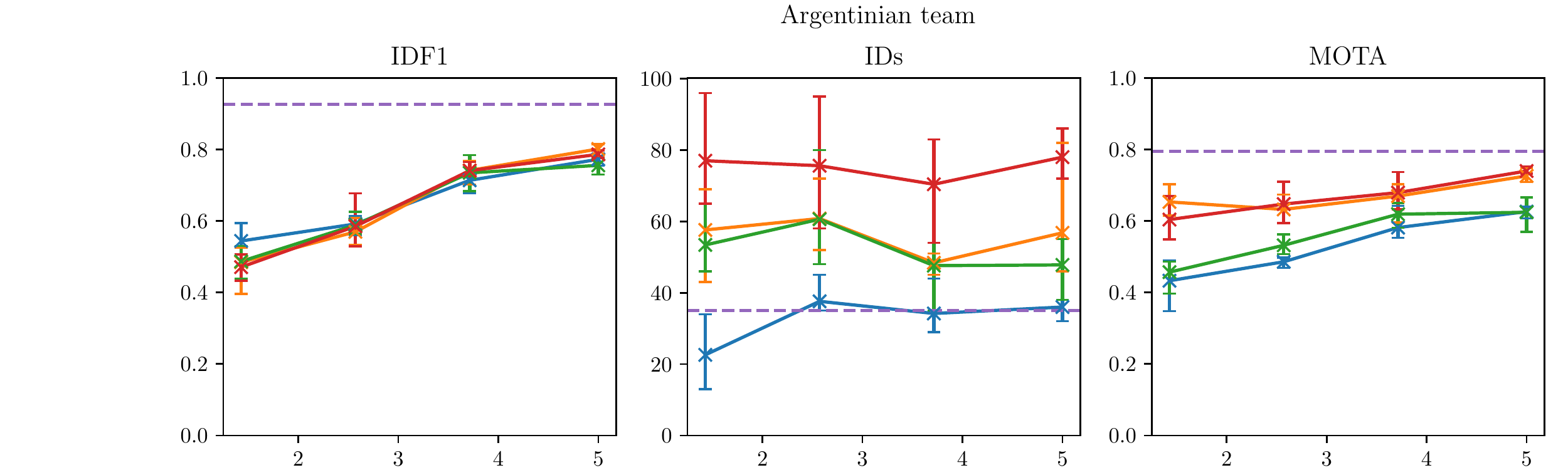} \\
\includegraphics[scale=\graphScale]{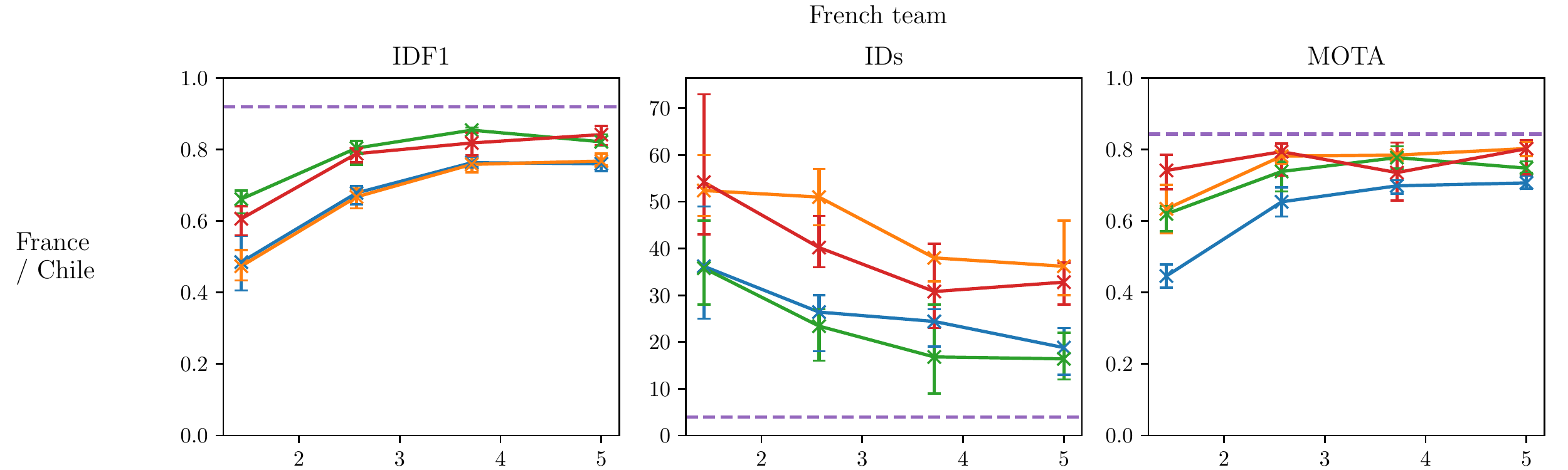} &
\includegraphics[scale=\graphScale]{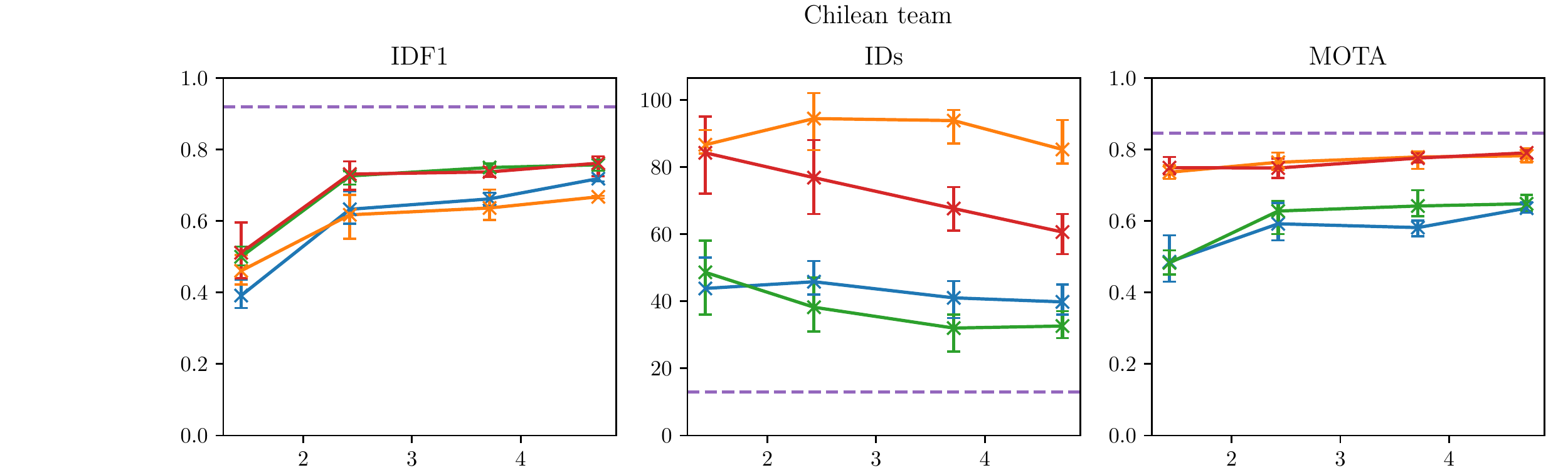} \\
\includegraphics[scale=\graphScale]{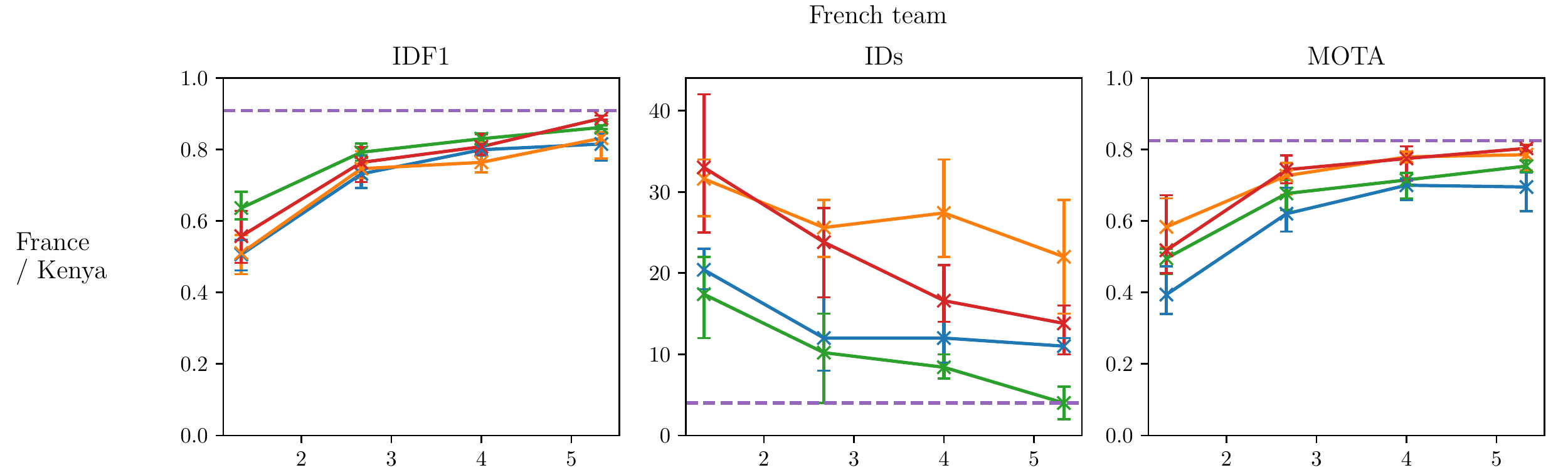} &
\includegraphics[scale=\graphScale]{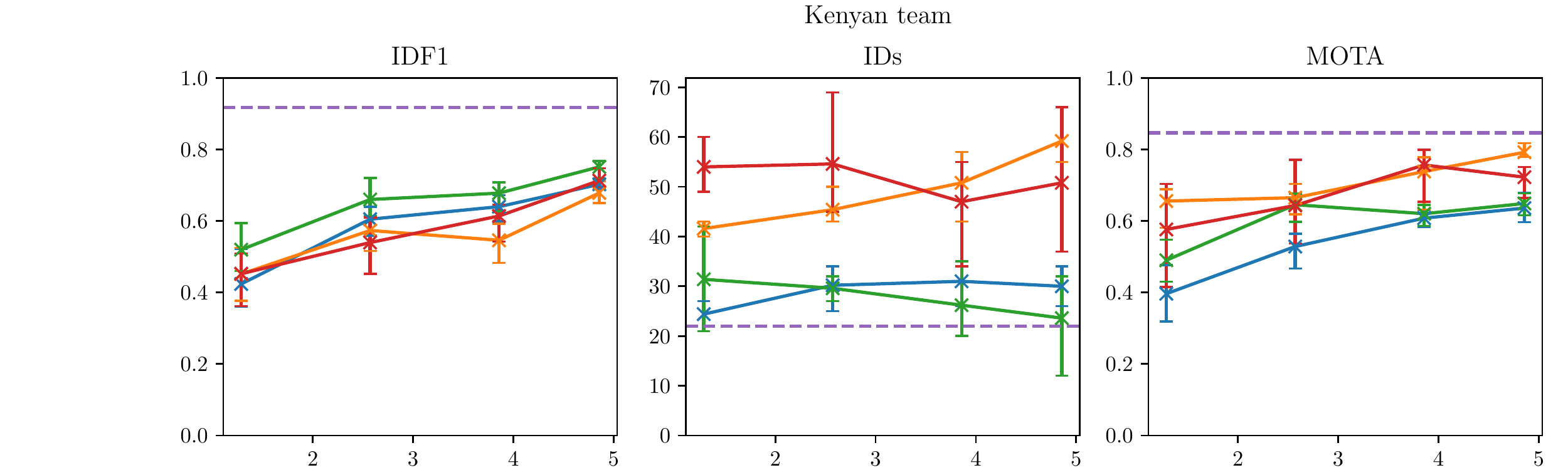} \\
\end{tabular}
\end{center}
   \caption{MOT metrics for the tracking of rugby sevens players in 3 videos. The x-axis corresponds to the total number of annotations divided by the number of tracked players.
   The variation intervals for the 5 seeds and average values are represented.
   The tested variants are:
   \(R_{img}\) frozen with the iterative association (\textcolor[RGB]{31,119,180}{\textbf{---}}),
   \(R_{img}\) frozen with the RNMF association (\textcolor[RGB]{255,127,14}{\textbf{---}}),
   \(R_{img}\) trained with the iterative association (\textcolor[RGB]{44,160,44}{\textbf{---}}),
   \(R_{img}\) trained with the RNMF association (\textcolor[RGB]{214,39,40}{\textbf{---}})
   and the ground truth association (\textcolor[RGB]{148,103,189}{\textbf{- - -}}).
   }
\label{fig:graph}
\end{figure*}

\subsubsection{Dataset}

Rugby sevens is a variant of rugby where two teams of seven players play a game composed of two seven minute halves. It is an Olympic sport since 2016.
We annotated a total of 58193 person bounding boxes in the images of three rugby sevens samples of 40 seconds to use them as ground truth for players of both teams, the referees and some people in the public.
These samples come from the Argentina / France, France / Chile and France / Kenya games of the 2021 Dubai tournament. They are encoded at a resolution of 1920 by 1080 pixels and a frame rate of 50 frames per seconds.
The aim of our experiments is to track players from one of the two teams taking part to the game.

Tracklets were extracted with the method detailed in section \ref{sec:tracklet_generation}. About 30\% of the tracklets have a number of frames superior to \(l_{min} = 10\). This represent an average of 346 tracklets per video of 40 seconds. These tracklets have an average length of one second and correspond to about 89\% of the detected bounding boxes.
We publicly release the tracking ground truth and the generated tracklets at \url{https://kalisteo.cea.fr/index.php/free-resources/}.

\subsubsection{Quantitative results and ablation studies}
\label{sec:ablation}

The annotator selects a number of tracklet examples for each player appearing in the sequence and also for the class 0 (opponent team, referees, public). At each round of annotations, a new user annotation for each player and two user annotations for the class 0 are added on average. As the training of our system is quick, the user can observe the consequences of the added annotations on the classification results and correct the big mistakes for the next round of annotations (for example false positives with high scores).

Once the user annotations have been added, we train the network with the same user annotations and 5 different seeds. We then compute standard MOT metrics \cite{ristani2016performance}: IDF1, MOTA and ID switches. Since our main objective is to correctly identify each player, the IDF1 metric is the most important to observe. MOTA is however key to report the completeness of the tracking bounding boxes for each player.
Figure \ref{fig:graph} shows the results of our method obtained with four variants. Results are analyzed according to two conditions: \(R_{img}\) frozen or trained and with the iterative association algorithm or with the RNMF algorithm.
As small tracklets are filtered, our method cannot achieve 100\% performance. In order to estimate the upper performance limit, we associate each tracklet to the ground truth. However, since our generated tracklets are not perfect, their association to the ground truth may also be ambiguous, which explains the not null ID switch limits.

\begin{figure}
\begin{center}
\includegraphics{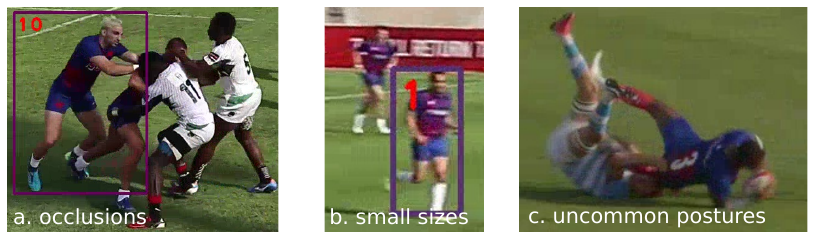}
\end{center}
   \caption{Illustration of complex situations that lead to missing detections and identification of players (here the players in blue).}
\label{fig:errors}
\end{figure}

\textbf{Number of annotations}. For the three video extracts, the more user annotations are provided, the best the MOT metrics are. However, we can observe that above the third round of annotations (about 3.5 annotations per player), the metrics only slightly improve and sometimes slightly deteriorate. This performance threshold can be explained by the difficult tracking conditions of some instants: the players are sometimes highly occluded or very small, there are very few details to identify them and the detection is difficult on complex postures. Some errors are illustrated on Figure \ref{fig:errors}. From the first to the third round of annotations, with \(R_{img}\) frozen and the iterative association algorithm, IDF1 and MOTA metrics increases on average respectively by 11 and 9 p.p. (percentage points) while the ID switches is divided by 5.

\textbf{Association algorithm choice}. The global RNMF optimization matches an identity to each tracklet but sometimes generates conflicts and wrong associations. This leads to better MOTA metrics as more detections are kept than with the simple iterative algorithm.
For the third round of annotations, the MOTA metric is increased by 12 p.p. on average when \(R_{img}\) is frozen.
However, the IDF1 metric is decreased by 1 p.p. and the number of ID switches increases by 25.
The iterative association should therefore be prefered to minimize wrong identity associations. The RNMF algorithm however leads to a more complete tracking.

\textbf{Training strategy}. Our experiments demonstrate that, even if \(R_{img}\) is not fine-tuned with data from the target domain (\(R_{img}\) frozen), it is still able, thanks to the Transformer network, to generate relevant features to re-identify the players. For the third round of annotations, the IDF1 and MOTA metrics are respectively on average 75\% and 66\% with the iterative association algorithm.
The best results are however obtained when weights of \(R_{img}\) are also updated during training. For the third round of annotations, with the iterative association algorithm, the IDF1 and MOTA metrics are increased respectively by 3 and 2 p.p. The number of ID switches is reduced on average by 3.
When the weights of \(R_{img}\) are updated, the training time for the 120 epochs significantly increases (from 28 seconds to 25 minutes for 32 annotations) and the system is no longer interactive. Indeed, the number of trainable parameters raises from about 4 to 25 millions. So, the optimal usage is to create the user annotations with \(R_{img}\) frozen and once the user is satisfied with the results, restart the training with the same annotations and \(R_{img}\) updated to obtain even better results.

\subsubsection{Comparison with state of the art multiple person tracking methods}
\label{comparison_generic_tracking}

\begin{table}
    \begin{center}
    \footnotesize{
    \begin{tabular}{|c|c|c|c|c|c|}
      \hline
      Video & Method & IDF1 & IDs & MOTA \\
      \hhline{=====}
         & ByteTrack \cite{zhang2021bytetrack} & 48.8 & 26 & 49.4 \\
         Argentina & TWBW \cite{bergmann2019tracking} & 24.4 & 64 & 40.8 \\
         / France & MOT neur. solv. \cite{braso2020learning} & 34.0 & 54 & 33.9 \\
         & Ours & \textbf{76.8} & \textbf{17} & \textbf{64.6} \\
      \hline
         & ByteTrack \cite{zhang2021bytetrack} & 54.9 & 23 & 64.4 \\
         France & TWBW \cite{bergmann2019tracking} & 22.7 & 74 & 28.4 \\
         / Chile & MOT neur. solv. \cite{braso2020learning} & 29.6 & 53 & 40.2 \\
         & Ours & \textbf{84.3} & \textbf{21} & \textbf{75.4} \\
      \hline
         & ByteTrack \cite{zhang2021bytetrack} & 60.3 & 14 & 64.0 \\
         France & TWBW \cite{bergmann2019tracking} & 30.6 & 44 & 45.0 \\
         / Kenya & MOT neur. solv. \cite{braso2020learning} & 48.0 & 26 & 61.1 \\
         & Ours & \textbf{82.2} & \textbf{7} & \textbf{70.1} \\
      \hline
    \end{tabular}
    }
    \end{center}
    \caption{MOT metrics for the tracking of the rugby sevens French team players.}
    \label{table:track_table}
\end{table}

Generic tracking algorithms track all the persons appearing in the video frames. This would include in our case, players of both teams, the referees and the public. Our approach however track players from a single team. This makes the comparison not straightforward. Some approaches have been proposed for the tracking of team sport players with single moving views \cite{lu2013learning, hurault2020self} but the comparison is still not easy since their evaluation datasets are private.
We therefore decided to run generic tracking algorithms on our rugby sevens extracts. In order to make a fair comparison with our approach, we manually selected the tracks generated by these algorithms that are associated, even partially, with players from the French team.
We tested two online methods, TWBW tracker \cite{bergmann2019tracking} and ByteTrack \cite{zhang2021bytetrack}, with their detections. ByteTrack achieves a very high performance on the MOT 2017 challenge \cite{milan2016mot}. We also tested an offline method, the MOT neural solver \cite{braso2020learning}, with our detections.
The results are presented in Table \ref{table:track_table}. With the limitations mentioned above, the metrics shows significantly lower performances for generic trackers. This is probably due to difficulties to handle correctly the occlusions and the players entering or leaving the view field. It therefore justifies our usage of a closed identity gallery with few annotations to learn the player appearances. Compared to ByteTrack \cite{zhang2021bytetrack}, the IDF1 metric is increased on average by 26 p.p.

\subsection{Evaluation of player identification on a full rugby sevens game}

Our system aims to track and identify players on a full game. Yet, a human-annotated tracking ground truth for a full game would be costly to generate. We therefore decided to evaluate the detection and re-ID performance of our approach on 32 frames regularly sampled in the France / Kenya game and focus on the French players. With players changes, 12 French players in total participated to this game. The ground truth represents 128 players bounding boxes.
For each experiment, we trained the model with 5 different seeds using the same 70 annotations (about 6 per player).
Results are shown in Table \ref{table:classification_table}.
The best total detection and identification performance (53.6\%) is obtained when the \(R_{img}\) is trained and the RNMF association algorithm is used.
A significant number of French players are not detected or correctly identified. This happens when the players on the back are only visible on few pixels or when some players occlude others.
Nevertheless, the total recall goes up to 67.9\% for the bounding boxes with an area superior to the average area of all the ground truth bounding boxes (25214 pixels). This demonstrates that when the players are sufficiently visible, our system is able to track them during a full match with few annotations.

\begin{table}
    \begin{center}
    \footnotesize{
    \begin{tabular}{|c|c||c|c|c||c|}
      \hline
        \multirow{2}{*}{\(R_{img}\)}
        & \multirow{2}{*}{assoc.}
        & Det.
        & Team class.
        & Id. class.
        & Total \\
        
        &
        & recall
        & recall
        & recall
        & recall \\
      \hline
      \multicolumn{6}{l}{All detected bounding boxes} \\
      \hline
         frozen & iter. & \multirow{4}{*}{75.8} & 58.4\(\pm\)2.1 & 73.8\(\pm\)4.5 & 32.7\(\pm\)2.4 \\
      \cline{1-2}
      \cline{4-6}
        frozen & RNMF & & 74.6\(\pm\)2.5 & 60.9\(\pm\)6.5 & 34.5\(\pm\)4.6 \\
      \cline{1-2}
      \cline{4-6}
        trained & iter. & & 75.9\(\pm\)3.9 & \textbf{84.0\(\pm\)3.4} & 48.3\(\pm\)3.0 \\
      \cline{1-2}
      \cline{4-6}
        trained & RNMF & & \textbf{89.1\(\pm\)2.0} & 79.4\(\pm\)2.6 & \textbf{53.6\(\pm\)1.8} \\
      \hline
      \multicolumn{6}{l}{Big detected bounding boxes (area superior to 25214 pixels)} \\
      \hline
         frozen & iter. & \multirow{4}{*}{89.7} & 60.8\(\pm\)2.2 & 77.3\(\pm\)6.8 & 42.1\(\pm\)3.1 \\
      \cline{1-2}
      \cline{4-6}
        frozen & RNMF & & 72.3\(\pm\)2.2 & 66.4\(\pm\)5.0 & 43.1\(\pm\)4.4 \\
      \cline{1-2}
      \cline{4-6}
        trained & iter. & & 76.2\(\pm\)3.5 & \textbf{87.4\(\pm\)5.2} & 59.7\(\pm\)4.2 \\
      \cline{1-2}
      \cline{4-6}
        trained & RNMF & & \textbf{90.8\(\pm\)0.9} & 83.5\(\pm\)3.4 & \textbf{67.9\(\pm\)2.6} \\
      \hline
    \end{tabular}
    }
    \end{center}
    \caption{French player detection and classification results on 32 frames of the France / Kenya game for 5 different seeds. Average values and standard deviations are provided. The detection recall corresponds to the number of players detected. The team classification recall corresponds to the number of players classified as French among the detected players. The identity classification recall corresponds to the number of correctly identified players among the players classified as French. The total recall is the product of all the previous columns and represents the complete performance of our system.}
    \label{table:classification_table}
\end{table}

\section{Conclusion}

In this paper, we proposed a new method to track team sport players with few user annotations. We demonstrated the performance of our approach on a rugby sevens dataset that we publicly release. We also showed that our method can track rugby sevens players during a full match with the annotation of only 6 few seconds length tracklets per player if they are observable with a minimal resolution. To our knowledge, no previous work on tracking of rugby players has been published. As future work, we would like to improve the detection of small and partially occluded players. Since our approach can be applied to any team sport, we would like to test it on other sports such as basketball. We also believe that the user annotation step would be sped up if an active learning process could smartly suggests tracklets to annotate.

\section{Acknowledgments}

This work benefited from a government grant managed by the French National Research Agency under the future investment program (ANR-19-STHP-0006) and the FactoryIA supercomputer financially supported by the Ile-de-France Regional Council.
The videos of our Rugby Sevens dataset are the courtesy of World Rugby. We would also like to thank Jérôme Daret, Jean-Baptiste Pascal and Julien Piscione from the French Rugby Federation for making this work possible.

{\small
\bibliographystyle{ieee_fullname}
\bibliography{trackMerger}
}

\end{document}